\documentclass{article}

\usepackage{PRIMEarxiv}

\usepackage[utf8]{inputenc} 
\usepackage[T1]{fontenc}    
\usepackage{hyperref}       
\usepackage{url}            
\usepackage{multirow}       
\usepackage{amsmath}        
\usepackage{tabularx}       
\usepackage{booktabs}       
\usepackage{amsfonts}       
\usepackage{nicefrac}       
\usepackage{microtype}      
\usepackage{svg}            
\usepackage{fancyhdr}       
\usepackage{graphicx}       
\graphicspath{{media/}}     

\title{EndoDINO: A Foundation Model for GI Endoscopy
}

\author{
  Patrick Dermyer\\
  Virgo Surgical Video Solutions, Inc.\\
  Denver, CO, USA\\
  patrick@virgosvs.com\\
  \And
  Angad Kalra\\
  Virgo Surgical Video Solutions, Inc.\\
  Vancouver, BC, Canada\\
  angad@virgosvs.com\\
  \And
  Matt Schwartz\\
  Virgo Surgical Video Solutions, Inc.\\
  Carlsbad, CA, USA\\
  matt@virgosvs.com\\
}

\begin{document}
\maketitle

\begingroup
\hyphenpenalty=10000

\begin{abstract}

In this work, we present EndoDINO, a foundation model for GI endoscopy tasks that achieves strong generalizability by pre-training on a well-curated image dataset sampled from the largest known GI endoscopy video dataset in the literature. Specifically, we pre-trained ViT models with 1B, 307M, and 86M parameters using datasets ranging from 100K to 10M curated images. Using EndoDINO as a frozen feature encoder, we achieved state-of-the-art performance in anatomical landmark classification, polyp segmentation, and Mayo endoscopic scoring (MES) for ulcerative colitis with only simple decoder heads.

\end{abstract}

\endgroup


\section{Introduction}
Due to their usefulness in a wide range of diagnostic and therapeutic applications, it is estimated that more than 20 million gastrointestinal (GI) endoscopies are performed in the United States each year \cite{ruhl2008niddkendoscopy}. Owing to the fact that GI endoscopy is both operator-dependent and generates a significant amount of high-dimensional video data, it has proven to be a fruitful area for AI research. As a result, GI endoscopy leads all medical specialties in the number of randomized controlled trials published on the use of AI clinical decision support tools \cite{han2023rctai}.

Applications of AI in GI endoscopy include tasks such as computer-assisted detection (CADe), computer-assisted diagnosis (CADx), and disease severity scoring for conditions including colorectal polyps, Barrett's esophagus, gastric cancer, and inflammatory bowel disease (IBD) \cite{Okagawa2022AIinendoscopy}. Further efforts have been made to leverage AI for procedural quality assessment and identification of anatomical landmarks \cite{Thakkar2020livecolonquality}. Monocular simultaneous localization and mapping (SLAM) in endoscopy holds promise for mapping the portion of the GI tract that is examined or potentially even automating endoscopic procedures \cite{ozyoruk2020endoslamdatasetunsupervisedmonocular}.

Research and development of AI for GI endoscopy has been hampered by the lack of large and diverse endoscopic datasets. As a result, traditional development in the field involves the use of natural image datasets such as ImageNet-1K for pre-training or transfer learning. This path requires the assembly of relatively large and well-labeled fine-tuning datasets to achieve meaningful results. Creation of such datasets is time-consuming and costly. Furthermore, the lack of a foundational backbone model means different AI solutions in GI endoscopy require their own inference compute overhead. From a deployment perspective, this limits the practicality of deploying multiple AI solutions to run in real-time during a procedure. A vision foundation model specifically pre-trained for GI endoscopy could unify AI solutions, accelerate research, and even unlock new capabilities to advance patient care.

\subsection{Related Work}
\paragraph{Natural Images}In computer vision for natural images, self-supervised pre-training on massive and well-curated datasets has been shown to produce powerful backbone features that achieve state-of-the-art performance on most vision tasks, ranging from classification to segmentation to depth estimation \cite{oquab2024dinov2learningrobustvisual}. Natural images benefit from high-quality open datasets, most notably ImageNet-1K \cite{deng2009imagenet}. The DINOv2 family of models extended this concept with the curation of the LVD-142M dataset. The curation process for LVD-142M involved deduplication and self-supervised image retrieval such that a starting set of 1.2B initial images was reduced to a curated set of 142M images \cite{oquab2024dinov2learningrobustvisual}.

\paragraph{Radiology}Building on the DINOv2 methodology, several groups have shown that the same self-supervised learning techniques that work well in the domain of natural images also work well in specific narrow domains such as medical imaging. Both RayDINO \cite{moutakanni2024advancinghumancentricairobust} and RAD-DINO \cite{pérezgarcía2024raddinoexploringscalablemedical} achieved state-of-the-art results on a range of chest X-ray evaluations by pre-training on large, unlabeled chest X-ray datasets and adapting small task-specific model heads. These efforts demonstrated that both pre-training from randomly initialized weights and continuation training from model weights derived from natural image pre-training produce strong performance on downstream tasks.

\paragraph{Pathology}These findings were replicated and extended in pathology. The models Virchow2 \cite{zimmermann2024virchow2scalingselfsupervisedmixed} and PLUTO \cite{juyal2024plutopathologyuniversaltransformer} utilized datasets of 3.1M and 160K whole slide images respectively in self-supervised pre-training architectures that were similar to DINOv2. Despite the similarity to DINOv2, both Virchow2 and PLUTO did introduce novel architectural decisions related to the domain-specific adaptation, such as multi-magnification to better capture concepts at the cellular, tissue, and whole slide level. These models now represent state-of-the-art performance in a wide range of pathology tasks.

\paragraph{GI Endoscopy}Similar attempts have been made to extend self-supervised pre-training to the domain of GI endoscopy. The largest and most popular open dataset of GI endoscopy is called HyperKvasir, which consists of 10,662 labeled images and 99,417 unlabeled images \cite{Borgli2020}. The labeled images are categorized into 23 different classes that include anatomical landmarks, MES, and a subset of polyp images with segmentation masks. Initial efforts to develop self-supervised pre-trained feature extractors for GI endoscopy leveraged the HyperKvasir dataset with training architectures such as MoCo v3, Barlow Twins, and DINOv1 \cite{sanderson2024studyselfsupervisedpretrainingvision, yao2023unsupervisedsegmentationcolonoscopyimages}. These efforts showed promise and were soon followed by the utilization of private and public/private datasets.

Leveraging data from 5 clinical trials of Etrolizumab, Yao et al. conducted DINOv1 self-supervised pre-training using data from 5,145 colonoscopy and sigmoidoscopy videos from patients with moderate to severe ulcerative colitis \cite{yao2023unsupervisedsegmentationcolonoscopyimages}. Other work by Chaitanya et al. conducted DINOv2 self-supervised pre-training using data from 4,911 colonoscopy and sigmoidoscopy videos from both ulcerative colitis and Crohn's disease clinical trials \cite{chaitanya2024argesspatiotemporaltransformerulcerative}. These models all demonstrated meaningful capabilities specifically for IBD-related tasks; however, they did not demonstrate state-of-the-art generalizability for non-IBD tasks. Another noteworthy effort made by Wang et al. utilized a combination of public datasets consisting of over 33K video clips and 5M frames to specifically study the impact of self-supervised pre-training on polyp classification, detection, and segmentation tasks \cite{wang2024foundationmodelendoscopyvideo}.

\subsection{Contributions}
This paper presents a study on the application of a significantly larger pre-training dataset to learn visual features in a self-supervised fashion that are truly generalizable to a wide range of GI endoscopy tasks. We studied three different model sizes (ViT-B, ViT-L, and ViT-g) \cite{dosovitskiy2021imageworth16x16words, zhai2022scalingvisiontransformers} and utilized data curation strategies to test different pre-training dataset sizes. Whereas prior related work has conducted evaluations using holdout data that is highly correlated with the pre-training data (e.g. HyperKvasir unlabeled vs. labeled images or holdout data from the same clinical trial or family of trials) we evaluated our pre-trained backbone models on evaluation data that has no direct relation to our pre-training data. Evaluation tasks included anatomical landmark classification, polyp segmentation, and Mayo endoscopic scoring, which we believe demonstrate superior generalizability over prior work.

Table \ref{tab:ssldatasets} summarizes the data and training architectures used in prior work on self-supervised learning (SSL) in GI endoscopy relative to EndoDINO.

\renewcommand{\arraystretch}{1.4}
\begin{table}[ht]
\caption{Overview of Prior SSL Datasets Relative to EndoDINO}
\centering
\begin{tabularx}{\textwidth}{lXXXX}
    \toprule
         & \textbf{Etrolizumab} & \textbf{EndoFM} & \textbf{ArgesFM} & \textbf{EndoDINO} \\
    \midrule
    \textbf{Citation}    & Yao et al.\cite{yao2023unsupervisedsegmentationcolonoscopyimages}
                         & Wang et al.\cite{wang2024foundationmodelendoscopyvideo}
                         & Chaitanya et al.\cite{chaitanya2024argesspatiotemporaltransformerulcerative}
                         & Ours \\
    \textbf{Total Video Clips}    & 5,145    & 33,000    & 3,927    & 130,037 \\
    \textbf{Total Images}         & \(\sim5\text{M}-13\text{M}\)   & 5M    & 61M    & 3.5B \\
    \textbf{Curation Methods}   & 1 fps sampling and low-quality frame removal \cite{becker2021framequality}
                                  & N/A
                                  & N/A
                                  & 5fps sampling, de-duplication, and hierarchical k-means clustering \\
    \textbf{Images Used}          & 525,711    & 5M    & 61M    & Up to 10M \\
    \bottomrule
\end{tabularx}
\label{tab:ssldatasets}
\end{table}
\renewcommand{\arraystretch}{1.0}

\section{Data and Methods}
\label{sec:headings}

We assembled our dataset by first retrieving still frames from a pool of 130,037 de-identified endoscopy videos, totaling over 3.5 billion still frames. All videos were captured using the VirgoCloud platform from Virgo Surgical Video Solutions, Inc. Videos were randomly selected to create a set that is approximately representative of upper and lower GI endoscopy in the United States. This set is inclusive of videos from a wide range of procedures, indications, and endoscopy equipment.

\subsection{Data Curation}

From the initial set of 3.5 billion frames, we initially downsampled from 30 frames per second to 5 frames per second. Frame embeddings were extracted using a DINOv2 encoder. Duplicate and near-duplicate frames were removed to create an index set of 112 million frames. Prior work in other domains has shown that dataset curation can be automated to provide improved representative data distribution for self-supervised learning \cite{vo2024automaticdatacurationselfsupervised}. Following the procedure in Vo et al., we conducted balanced sampling from a hierarchical k-means clustering of the index frames, using 5m, 250k, 25k and 5k clusters in 4 levels. Datasets for pre-training, sized 100,000 through 10 million, were created by balanced sampling from these clusters. Subjective review showed the importance of near-duplicate removal prior to hierarchical clustering for clustering quality.

\subsection{Pre-training}

We conducted pre-training using the DINOv2 methodology and explored a range of hyperparameters including model size, dataset size, number of head prototypes, batch size, crop size, and learning rate. Pre-training was conducted on a cluster of 8 NVIDIA H100 GPUs for up to 625,000 iterations. Checkpoints were saved every 25,000 iterations for downstream task evaluation.

\begin{figure}[ht]
    \centering
    \includegraphics[width=0.7\textwidth]{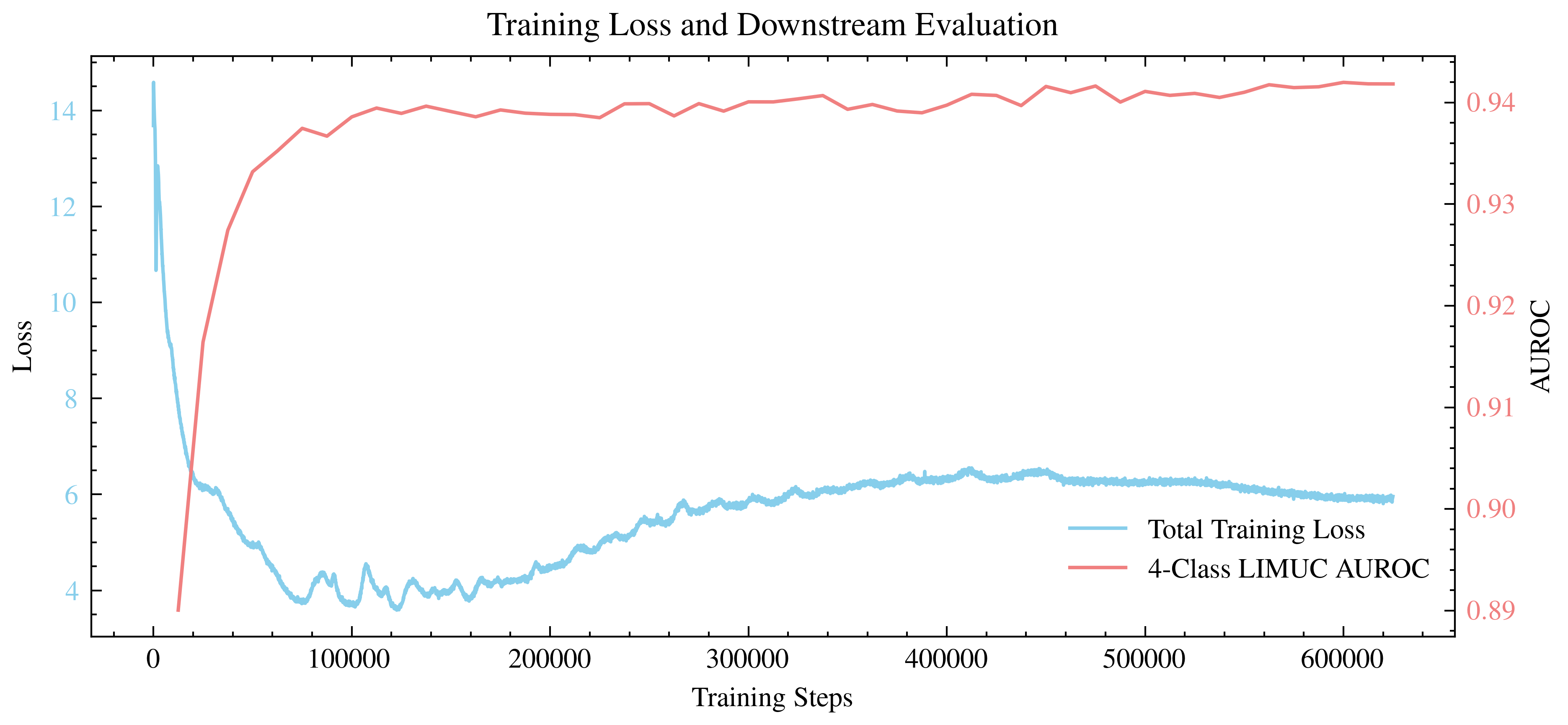}
    \caption{Total DINOv2 training loss in blue, compared to performance on downstream LIMUC 4 class MES task in red, by training step. Peak performance on this downstream task occurs long after total DINOv2 loss is minimized, highlighting the importance of selecting model checkpoints based on downstream task performance.}
    \label{fig:endodinoandlimuc}
\end{figure}

We noted that DINOv2 losses did not perfectly correspond to performance on various downstream tasks. As such, we created an evaluation pipeline on an array of downstream tasks to select our best-performing checkpoints. Figure \ref{fig:endodinoandlimuc} shows an overlay of DINOv2 loss with performance on our LIMUC 4 class MES task for our ViT-g/14 model.

\subsection{Evaluations}

\paragraph{HyperKvasir} The HyperKvasir dataset was used to generate three evaluation tasks that are standard in the literature. These tasks were anatomical landmark classification, three-class Mayo endoscopic scoring, and polyp segmentation. For each of these tasks, data was split 80/10/10 for fine-tuning, validation, and testing, respectively. For the anatomical landmark classification task, we sought to study the few-shot learning capabilities of our pre-trained models and thus split the data 1/10/10 for comparison to prior state-of-the-art.

\begin{figure}[ht]
    \centering
    \includegraphics[width=0.45\textwidth]{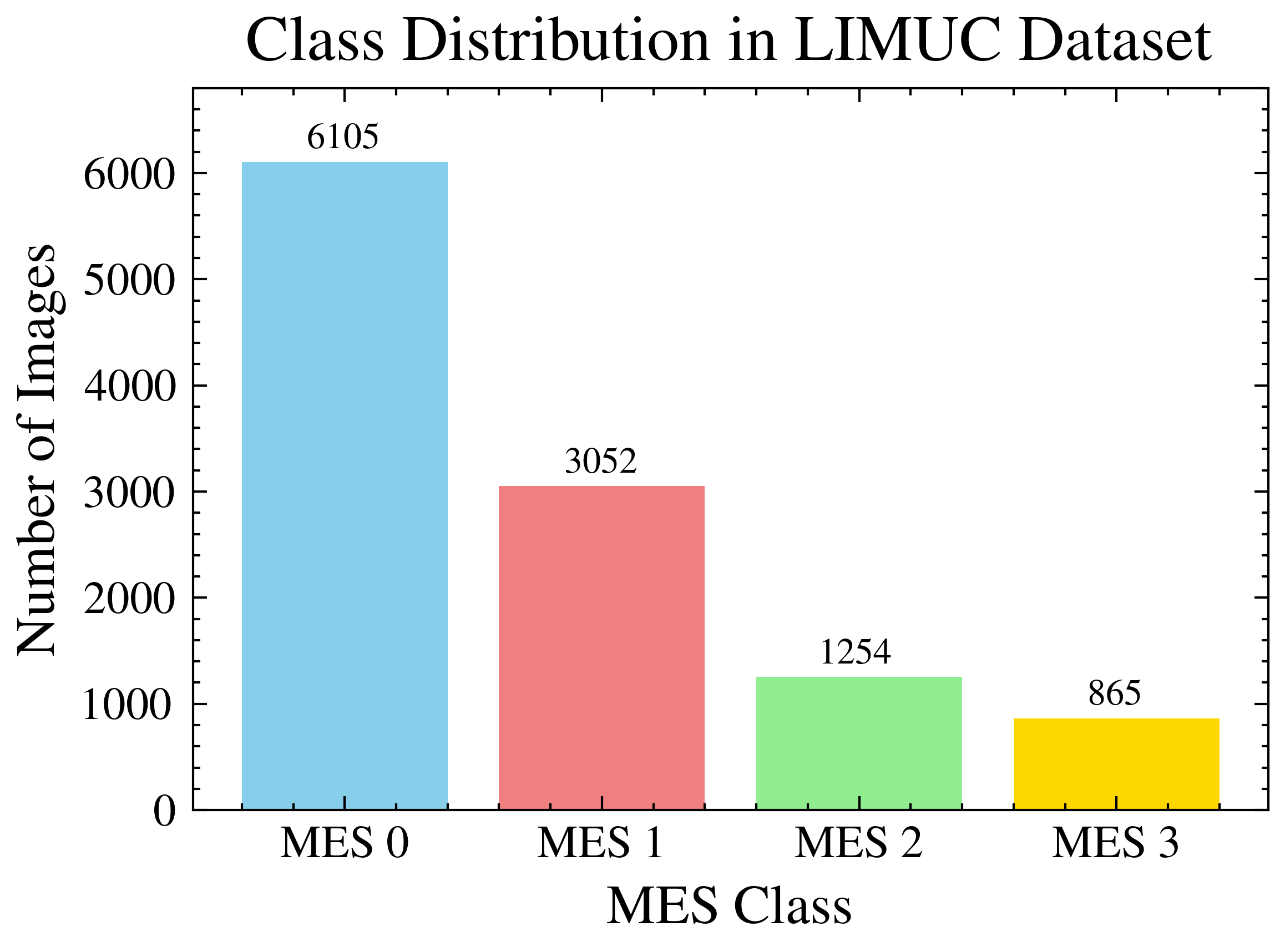}
    \caption{Representation of classes in the LIMUC dataset showing significant class imbalance that favors lower MES scores.}
    \label{fig:limuchistogram}
\end{figure}

\paragraph{LIMUC} A larger open dataset of labeled images from ulcerative colitis endoscopy videos, called Labeled Images for Ulcerative Colitis (LIMUC), was released in 2022 \cite{polat2022mayolimuc}. This set consists of 11,276 images that are labeled using the full 4-class Mayo endoscopic score. The distribution of videos by MES is shown in Figure \ref{fig:limuchistogram}. For evaluation purposes, we followed the methodology of the original paper, which called for a 15\% holdout test set, with the remaining 85\% used for 10-fold cross-validation.

\section{Experiments}

In this section, we present the results of various evaluation experiments and place our results in comparison to prior work. Specifically, we studied anatomical landmark classification, polyp segmentation, 3-class Mayo endoscopic scoring, and 4-class Mayo endoscopic scoring. For each experiment, we describe the adaptations used with EndoDINO and present our results. It should be noted that our focus was demonstrating simplicity in the adaptations rather than maximizing performance on individual tasks.

\subsection{Anatomical Landmark Classification}
We compared our anatomical landmark classification performance to that of Sanderson et al. using their 80/10/10 data split of the labeled HyperKvasir dataset. This included anatomical landmarks from both upper and lower endoscopy. Additionally, we test EndoDINO by downsampling the training set to 1\% and maintaining the same validation and test splits to demonstrate few-shot learning performance with at most 7 examples per class.

To evaluate our EndoDINO models, we used simple linear probing as in Oquab et al. 2024 without data augmentation. We report in Table \ref{tab:anatomical} the performance of LVD-142m DINOv2 models and our ViT-B/14 model on the full training set with the best results from Sanderson et al. We also report the performance of our models on the 1\% training set. The dramatic increase in macro F1 score on the 1\% models can be explained by the more balanced training subset. 

\begin{table}[ht]
\caption{Anatomical Landmark Classification}
\centering
\begin{tabular}{l l l l l l}
    \toprule
    & \textbf{Backbone} & \textbf{Pre-training Data} & \textbf{Pre-training Algorithm} & \textbf{Macro F1} & \textbf{Micro F1} \\
    \midrule
    \multirow{4}{*}{\textbf{Prior - 80/10/10 Split}} 
    & ResNet50 & ImageNet-1K  & MoCo v3          & 0.828 & 0.993 \\
    & ViT-B    & ImageNet-1K  & MoCo v3          & 0.828 & 0.993 \\
    & ViT-B/14 & LVD-142M     & DINOv2 w/reg     & 0.833 & 0.998 \\
    & ViT-g/14 & LVD-142M     & DINOv2 w/reg     & 0.833 & 0.998 \\
    \cmidrule{2-6}
    \multirow{1}{*}{\textbf{Ours - 80/10/10 Split}}
    & ViT-B/14 & EndoDINO-1M  & DINOv2 w/reg     & 0.833 & 0.998 \\
    \cmidrule{2-6}
    \multirow{3}{*}{\textbf{Ours - 1/10/10 Split}}
    & ViT-B/14 & EndoDINO-1M  & DINOv2 w/reg     & 0.911 & 0.990 \\
    & ViT-L/14 & EndoDINO-5M  & DINOv2 w/reg     & 0.997 & 0.998 \\
    & ViT-g/14 & EndoDINO-10M & DINOv2 w/reg     & 0.995 & 0.995 \\
    \bottomrule
\end{tabular}
\label{tab:anatomical}
\end{table}

\subsection{Polyp Segmentation}
To compare performance in polyp segmentation on the HyperKvasir data, we replicated the training procedure outlined by Sanderson et al. on our \textit{frozen} EndoDINO backbones using three different segmentation heads: Linear, Boosted Linear, and DPT \cite{ranftl2021vitdpt}. Note that Sanderson et al. performed end-to-end training, i.e. all model parameters are fine-tuned for all of their segmentation models. Our models used frozen backbone features and only trained the decoder head parameters. The Linear head input is the final patch tokens, whereas the Boosted Linear head input is the concatenation of the patch tokens of the last four backbone layers, as described in \cite{oquab2024dinov2learningrobustvisual}. Input image resolution, data split ratios (80/10/10), training data augmentations, and relevant hyper-parameter values are equivalent between our models and the best ViT-B/14 model from Sanderson et al.

\begin{figure}[ht]
    \centering
    \includegraphics[width=0.65\textwidth]{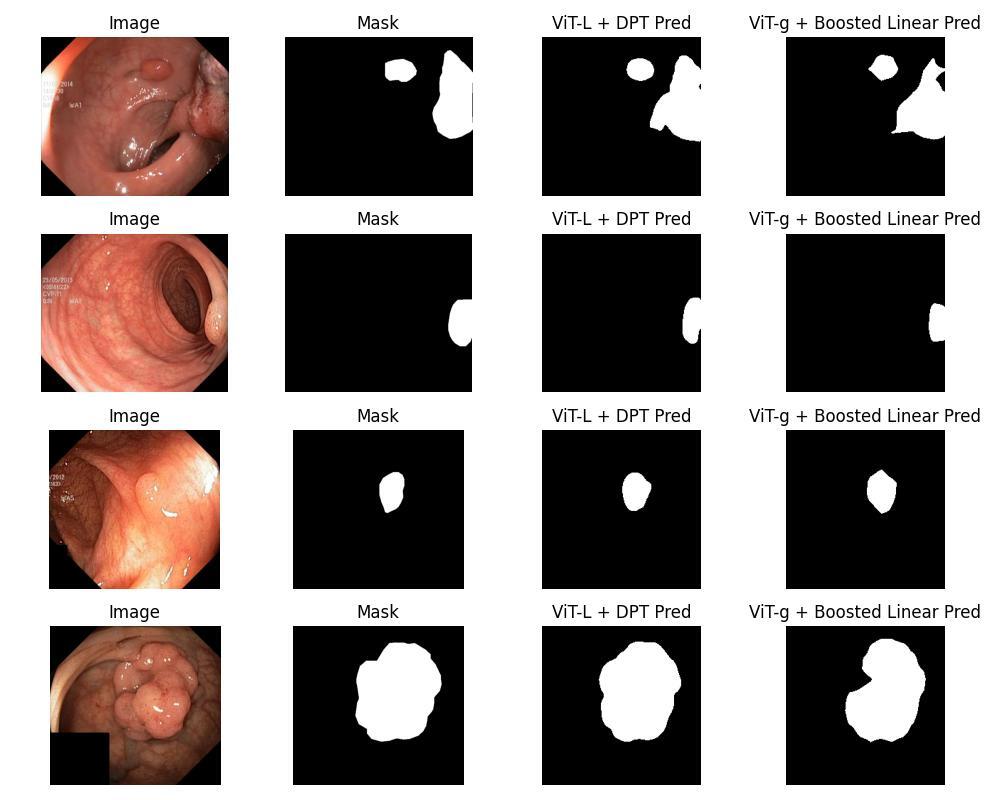}
    \caption{Representative examples of KvasirSEG polyp images alongside ground truth masks and our predictions for EndoDINO ViT-L/14 with DPT head and EndoDINO ViT-g/14 with Boosted Linear head models.}
    \label{fig:polypseg}
\end{figure}

Table \ref{tab:polypseg} shows the performance breakdown for the polyp segmentation models. Our frozen ViT-B/14 EndoDINO model achieves very similar results to the top performing model from Sanderson et al. without requiring end-to-end training. We were able to surpass prior performance by switching to our ViT-L/14 frozen backbone, with the most notable gain occurring in mIoU. Furthermore, our ViT-g/14 frozen backbone with linear heads (1k to 6k parameters) is almost able to achieve similar performance as the best end-to-end trained ViT-B/14 model without using a complex decoder head (22M parameters). This shows that the information captured in the ViT-g/14 EndoDINO features allows simple linear decoder heads to achieve performance comparable to fully trained ViT-B/14 segmentation models. Figure \ref{fig:polypseg} shows representative samples of predicted image masks for our EndoDINO ViT-L/14 with DPT head and EndoDINO ViT-g/14 with Boosted Linear head models.

\begin{table}[ht]
\caption{KvasirSEG Polyp Segmentation}
\centering
\begin{tabular}{lllllllll}
    \toprule
    & \textbf{Backbone} & \textbf{Head} & \textbf{Pre-train. Data} & \textbf{Pre-train. Alg.} & \textbf{mDice} & \textbf{mIoU} & \textbf{mPrec} & \textbf{mRec} \\
    \midrule
    \multirow{1}{*}{\textbf{Prior}}
    & ViT-B/14       & DPT        & ImageNet-1K & MAE & 0.896 & 0.834 & 0.921 & 0.902 \\
    \cmidrule{2-9}
    \multirow{4}{*}{\textbf{Ours}}
    & ViT-B/14 & DPT & EndoDINO-1M & DINOv2 w/reg & 0.895 & 0.832 & 0.910 & 0.908     \\
    & ViT-L/14 & DPT & EndoDINO-5M & DINOv2 w/reg & \textbf{0.909} & \textbf{0.864} & 0.914 & \textbf{0.917}     \\
    & ViT-g/14 & Linear & EndoDINO-10M & DINOv2 w/reg & 0.875 & 0.828 & 0.906 & 0.878  \\
    & ViT-g/14 & Boosted Linear & EndoDINO-10M & DINOv2 w/reg & 0.875 & 0.829 & \textbf{0.925} & 0.862  \\
    \bottomrule
\end{tabular}
\label{tab:polypseg}
\end{table}

\subsection{3-class Mayo Endoscopic Scoring}
As in Yao et al. \cite{yao2023unsupervisedsegmentationcolonoscopyimages}, we evaluated the performance of EndoDINO features on the downstream task of 3-class Mayo endoscopic scoring for ulcerative colitis still images from the HyperKvasir dataset. In this dataset, there are 201 images labeled MES 1, 441 images labeled MES 2, 143 images labeled MES 3, and 0 images labeled MES 0. We followed the published HyperKvasir data split for 2-fold cross validation \cite{Borgli2020}.

We present in Table \ref{tab:3classmayo} the average macro-F1 and micro-F1 from 2-fold cross validation from domain self-supervised learning (SSL) and fully supervised learning from Yao et al. alongside results from linear probing our EndoDINO models. Linear probing on both DINOv2 and EndoDINO models was performed without data augmentation. Best results are in bold.

\begin{table}[ht]
\caption{HyperKvasir 3 Class Mayo Endoscopic Scoring}
\centering
\begin{tabular}{l l l l l l}
    \toprule
    & \textbf{Backbone} & \textbf{Pre-training Data} & \textbf{Pre-training Algorithm} & \textbf{Macro F1} & \textbf{Micro F1} \\
    \midrule
    \multirow{5}{*}{\textbf{Prior}} 
    & ViT-B/16 & Etrolizumab  & DINOv1           & 0.706 & 0.739 \\
    & ViT-B/16 & EndoFM       & DINOv1           & 0.699 & 0.723 \\
    & DenseNet & N/A          & Fully Supervised & 0.729 & 0.751 \\
    & ViT-B/14 & LVD-142M     & DINOv2 w/reg     & 0.704 & 0.741 \\
    & ViT-g/14 & LVD-142M     & DINOv2 w/reg     & 0.735 & 0.758 \\
    \cmidrule{2-6}
    \multirow{3}{*}{\textbf{Ours}}
    & ViT-B/14 & EndoDINO-1M  & DINOv2 w/reg & 0.740 & 0.776 \\
    & ViT-L/14 & EndoDINO-5M  & DINOv2 w/reg & 0.740 & 0.768 \\
    & ViT-g/14 & EndoDINO-10M & DINOv2 w/reg & \textbf{0.748} & \textbf{0.779} \\
    \bottomrule
\end{tabular}
\label{tab:3classmayo}
\end{table}

\subsection{4-class Mayo Endoscopic Scoring}
As in Polat et al. \cite{polat2022mayolimuc}, we evaluated the performance of EndoDINO features on the downstream task of 4-class Mayo endoscopic scoring for ulcerative colitis. We followed their procedure of splitting 15\% of the images for the test set, while the remaining 85\% of images were used for 10-fold cross-validation. We adopt the same folds used in the original paper. For our best-performing model (EndoDINO ViT-g/14) we also train with a smaller 20\% subset of the training data in each fold.

We present in Table \ref{tab:4classmayo} the AUROC and macro F1 for all of our tested models, along with the macro F1 for the best model in Polat et al., as the prior work did not report an AUROC for tested models. We note that our models were trained without any data augmentation, whereas Polat et al. utilized augmentations such as horizontal flipping and random rotation and increased performance by using a regression-based approach. Best results are in bold.

\begin{table}[ht]
\caption{LIMUC 4 Class Mayo Endoscopic Scoring}
\centering
\begin{tabular}{l l l l l l}
\toprule
 & \textbf{Backbone} & \textbf{Pre-training Data} & \textbf{Pre-training Alg} & \textbf{AUROC} & \textbf{Macro F1} \\
\midrule
\multirow{3}{*}{\textbf{Prior}} 
 & DenseNet121 & ImageNet-1K & N/A & N/A & 0.697 \\
 & ViT-B/14    & LVD-142M    & DINOv2 w/reg & 0.926 & 0.680 \\
 & ViT-g/14    & LVD-142M    & DINOv2 w/reg & 0.927 & 0.681 \\
\midrule
\multirow{4}{*}{\textbf{Ours}}
 & ViT-B/14 & EndoDINO-1M-Random & DINOv2 w/reg & 0.935 & 0.695 \\
 & ViT-B/14 & EndoDINO-1M-HKM    & DINOv2 w/reg & 0.937 & 0.706 \\
 & ViT-L/14 & EndoDINO-5M-HKM    & DINOv2 w/reg & 0.940 & 0.713 \\
 & ViT-g/14 & EndoDINO-10M-HKM   & DINOv2 w/reg & \textbf{0.942} & \textbf{0.715} \\
\midrule
\textbf{Ours - 20\%}
 & ViT-g/14  & EndoDINO-10M-HKM   & DINOv2 w/reg & 0.933 & 0.689 \\
\bottomrule
\end{tabular}
\label{tab:4classmayo}
\end{table}

\section{Discussion and Future Work}
In this work, we present EndoDINO, a foundation model for GI endoscopy tasks pre-trained on a well-curated dataset of endoscopy images derived from the largest and most diverse set of endoscopy videos reported in the literature to date \cite{Zhu2023publicgidata}. Our experiments show state-of-the-art performance on a wide range of downstream GI endoscopy tasks and promising results on few-shot learning. Importantly, the EndoDINO encoders show strong performance generalizing to task-specific evaluation datasets derived from completely unrelated data capture efforts, even with minimal task-specific feature engineering. Relative to prior efforts at developing a generalized pre-trained backbone model for endoscopy, we attribute the strong performance of EndoDINO to several key factors: (1) a significantly larger pool of data including diversity in locations, equipment, and indications; (2) automated data curation techniques that naturally preserve general endoscopy concepts like blurriness, specularity, and bowel preparation; and (3) advances in self-supervised learning techniques.

In our experience, EndoDINO works well as a backbone model for the development of downstream AI tasks in GI endoscopy. There are several practical benefits from such a generalized backbone model. Namely, utilizing a backbone model streamlines downstream research and development by both reducing the need for acquiring task-specific labeled data and reducing the computational burden for training task-specific models. Front-loading the computational burden into pre-training also supports real-time applications of AI in GI endoscopy. For such applications, a typical GI endoscopy suite will likely have access to limited compute, perhaps a single desktop-class GPU. With EndoDINO as a backbone model, a single forward-pass provides powerful representative features that can simultaneously be used for multiple downstream tasks. This opens up the potential to efficiently run multiple AI task models (e.g. polyp segmentation, polyp classification, depth estimation, and localization) in parallel during a GI endoscopy procedure.

In future work, we plan to scale up the amount of videos and still frames used by at least a factor of ten and run ablation experiments to study the scaling properties of the image dataset size. Additionally, we will study the effects of modifying certain aspects of the pre-training process. Lastly, we intend to study certain novel emergent capabilities of our pre-trained models that may assist with precision medicine.

\section*{Acknowledgments}
We would like to thank the creators of the HyperKvasir and LIMUC datasets for contributing open access datasets to the field, which made our evaluations possible. We would also like to thank the teams at Meta AI and Microsoft Health Futures for answering a number of our questions about DINOv2 training for specific domains.

\bibliographystyle{unsrt}  
\bibliography{references}

\end{document}